\documentclass[runningheads]{llncs}

\usepackage{graphicx}
\usepackage{amsmath,amssymb}
\usepackage{color}
\usepackage{mathrsfs}
\usepackage{listings}
\usepackage{cite}
\usepackage{subfig}

\newcommand{\figsz}{0.13}

\begin{document}

\title{MobileFace: 3D Face Reconstruction\\ with Efficient CNN Regression}

\titlerunning{MobileFace}

\author{Nikolai Chinaev\inst{1} \and
Alexander Chigorin\inst{1} \and
Ivan Laptev\inst{1,2}}

\authorrunning{N. Chinaev \and A. Chigorin \and I. Laptev}

\institute{VisionLabs, Amsterdam, The Netherlands\\
\email{\{n.chinaev, a.chigorin\}@visionlabs.ru}\\
\and
Inria, WILLOW, Departement d'Informatique de l'Ecole Normale Superieure, PSL Research University, ENS/INRIA/CNRS 
UMR 8548, Paris, France\\
\email{ivan.laptev@inria.fr}}

\maketitle

\begin{abstract}
Estimation of facial shapes plays a central role for face transfer and animation.
Accurate 3D face reconstruction, however, often deploys iterative and costly methods preventing real-time applications.
In this work we design a compact and fast CNN model enabling real-time face reconstruction on mobile devices.
For this purpose, we first study more traditional but slow morphable face models and use them to automatically annotate 
a large set of images for CNN training.
We then investigate a class of efficient MobileNet CNNs and adapt such models for the task of shape regression.
Our evaluation on three datasets demonstrates significant improvements in the speed and the size of our model while
maintaining state-of-the-art reconstruction accuracy.
\keywords{3d face reconstruction \and morphable model \and CNN}
\end{abstract}

\section{Introduction}
%
%
3D face reconstruction from monocular images is a long-standing goal in computer vision with applications in face 
recognition, film industry, animation and other areas. Earlier efforts date back to late nineties and introduce
morphable face models~\cite{BaselMM}. Traditional methods address this task with optimization-based techniques and 
analysis-through-synthesis methods~\cite{BaselFitting, Piotraschke2016Automated3F, Blanz2003, thies2016face, GarridoRigs}. More 
recently, reg\-ression-based methods started to emerge \cite{zhu2015discriminative, ZFace, 3DDFA, MOFA}. In particular, 
the task has seen an increasing interest from the CNN community over the past few years \cite{3DDFA, MOFA, VRN, 
SynthLearn, tran2017regressing}. However, the applicability of neural networks remains difficult due to the lack of large-scale training data.
Possible solutions include the use of synthetic data \cite{ZFace, SynthLearn}, incorporation of unsupervised 
training criteria \cite{MOFA}, or combination of both \cite{Richardson_2017_CVPR}. Another option is to produce 
semi-synthetic data by applying an optimization-based algorithm with proven accuracy to a database of faces 
\cite{3DDFA, VRN, tran2017regressing}.

Optimization-based methods for morphable model fitting vary in many respects. Some design choices include image 
formation model, regularization and optimization strategy. Another source of variation is the kind of face attributes 
being used. Traditional formulation employs face texture \cite{BaselMM}. It uses morphable model to generate a 
synthetic face image and optimizes for parameters that would minimize the difference between the synthetic image 
and the target. However, this formulation also relies on a sparse set of facial landmarks used for initialization. 
Earlier methods used manually annotated landmarks \cite{Blanz2003}. The user was required to annotate a few facial 
points by hand. Recent explosion of facial landmarking methods \cite{Kazemi, SDM, FAN, MDM} made this process automatic 
and the set of landmarks became richer. This posed the question if morphable model fitting could be done based purely 
on landmarks \cite{ABas}. It is especially desirable because algorithms based on landmarks are much faster and 
suitable for real-time performance while texture-based algorithms are quite slow (on the order of 1 minute per image).

Unfortunately existing literature reports only few quantitative evaluations of optimization-based fitting algorithms.
Some works assume that landmark-based fitting provides satisfactory accuracy 
\cite{HighFidelityNormalization, PerspectiveAware} while others demonstrate its limitations \cite{ABas, OcContours}. 
Some use texture-based algorithms at the cost of higher computational demands, but the advantage in accuracy is not 
quantified \cite{thies2016face, GarridoRigs, Saito2017PhotorealisticFT}.
The situation is further complicated by the lack of standard benchmarks with reliable ground truth and well-defined evaluation
procedures.

We implement a morphable model fitting algorithm and tune its parameters in two scenarios: relying solely on 
landmarks and using landmarks in combination with the texture. We test this algorithm on images from BU4DFE dataset 
\cite{BU4DFE} and demonstrate that incorporation of texture significantly improves the accuracy.

It is desirable to enjoy both the accuracy of texture-based reconstruction algorithms and the high processing speed
enabled by network-based methods. To this end, we use the fitting algorithm to process 300W database of faces 
\cite{300W} and train a neural network to predict facial geometry on the resulting semi-synthetic dataset. It is 
important to keep in mind that the applicability of the fitting algorithm is limited by the expressive power of the 
morphable model. In particular, it doesn't handle large occlusions and extreme lighting conditions very well. To rule 
the failures out, we visually inspect the processed dataset and delete failed examples. We compare our dataset with a 
similarly produced 300W-3D \cite{3DDFA} and show that our dataset allows to learn more accurate models. We make our 
dataset publicly available \footnote{\url{https://github.com/nchinaev/MobileFace}}.

An important consideration for CNN training is the loss function. Standard losses become problematic when predicting
parameters of morphable face models due to the different nature and scales of individual parameters.
To resolve this issue, the MSE loss needs to be reweighted and some ad-hoc weighting schemes have been used in the past~\cite{3DDFA}.
We present a loss function that accounts for individual contributions of morphable model parameters in a clear and 
intuitive manner by constructing a 3D model and directly comparing it to the ground truth in the 3D space and in the 2D 
projected space.

This work provides the following contributions:
(i) we evaluate variants of the fitting algorithm on a database of facial scans providing quantitative evidence of
    texture-based algorithms superiority;
(ii)  we train a MobileNet-based neural network that allows for fast facial shape reconstruction even on mobile devices;
(iii) we propose an intuitive loss function for CNN training;
(iv)  we make our evaluation code and datasets publicly available.

%
%
	\subsection{Related Work}
%
%
Algorithms for monocular 3d face shape reconstruction may be broadly classified into two following categories: 
optimization-based and regression-based. Optimization-based approaches make assumptions about the nature of image 
formation and express them in the form of energy functions. This is possible because faces represent a set of objects 
that one can collect some strong priors about. One popular form of such prior is a morphable model. Another way to 
model image formation is shape from shading technique \cite{Kemelmacher1, Kemelmacher2, roth2016adaptive}. This 
class of algorithms has a drawback of high computational complexity. Regression-based methods learn from 
data. The absence of large datasets for this task is a limitation that can be addressed in several ways outlined below.
\medskip\\
\noindent
\textbf{Learning From Synthetic Data.} Synthetic data may be produced by rendering facial scans \cite{ZFace} or by
rendering images from a morphable model \cite{SynthLearn}. Corresponding ground truth 3d models are readily available 
in this case because they were used for rendering. These approaches have two limitations: first, the variability in 
facial shapes is only limited to the subjects participating in acquisition, and second, the image formation is limited 
by the exact illumination model used for rendering.
\medskip\\
\noindent
\textbf{Unsupervised Learning}. Tewari et al. \cite{MOFA} incorporate rendering process into their learning framework. 
This rendering layer is implemented in a way that it can be back-propagated through. This allows to circumvent the 
necessity of having ground truth 3d models for images and makes it possible to learn from datasets containing face 
images alone. In the follow up work Tewari et al. \cite{MOFACooler} go further and learn corrections to the morphable 
model. Richardson et al. \cite{Richardson_2017_CVPR} incorporate shape from shading into learning process to learn 
finer details.
\medskip\\
\noindent
\textbf{Fitting + Learning.} Most closely related to our work are works of Zhu et al. \cite{3DDFA} and Tran et al. 
\cite{tran2017regressing}. They both use fitting algorithms to generate datasets for neural network training. However,
accuracies of the respective fitting algorithms \cite{BaselFitting} and \cite{Piotraschke2016Automated3F} in the 
context of evaluation on datasets of facial scans are not reported by their authors. This raises two questions: what is 
the maximum accuracy attainable by learning from the results of these fitting methods and what are the gaps between the 
fitting methods and the respective learned networks? We evaluate accuracies of our fitting methods and networks on 
images from BU4DFE dataset in our work.
%
%
\section{MobileFace}
%
%
Our main objective is to create fast and compact face shape predictor suitable for real-time inference on mobile 
devices. To achieve this goal we train a network to predict morphable model parameters (to be introduced in 
Sec. \ref{sec:MM}). Those include parameters related to 3d shape $\vec{\alpha_{id}}$ and 
$\vec{\alpha_{exp}}$, as well as those related to projection of the model from 3d space to the image plane: translation 
$\vec{t}$, three angles $\phi$, $\gamma$, $\theta$ and projection $f$, $P_x$, $P_y$. Vector $\vec{p} \in 
\mathbb{R}^{118}$ is a concatenation of all the morphable model parameters predicted by the network:
\begin{align}\label{eq:vectorP}
\vec{p} = \left(\begin{array}{ccccccccc} \vec{\alpha_{id}}^T & ~~\vec{\alpha_{exp}}^T & ~~\vec{t}^T & ~~\phi & ~~\gamma 
& ~~\theta & ~~f & ~~Px & ~~Py \end{array} \right)^T
\end{align}
%
%
	\subsection{Loss Functions}
%
%
We experiment with two losses in this work. The first MSE loss can be defined as
\begin{align}\label{eq:MSEloss}
\texttt{Loss}_{\texttt{MSE}} = \sum_{i}{||\vec{p^i} - \vec{p^i}_{gt}||_2^2}.
\end{align}
Such a loss, however, is likely to be sub-optimal as it treats parameters $\vec{p}$ of different nature and scales equally. They impact the $3d$ 
reconstruction accuracy and the projection accuracy differently. One way to overcome this is to use the outputs of the 
network to construct 3d meshes $\vec{S}(\vec{p^i})$ and compare them with ground truth $\vec{S}_{gt}$ during training 
\cite{dou2017end}. However, such a loss alone would only allow to learn parameters related to the 3d shape: 
$\vec{\alpha_{id}}$ and $\vec{\alpha_{exp}}$. To allow the network to learn other parameters, we propose to augment 
this loss by an additional term on model projections $P(\vec{p^i})$:
\begin{align}\label{eq:loss}
\texttt{Loss}_{\texttt{2d + 3d},~l_2} = \sum_{i}{\left(||\vec{S}(\vec{p^i}) -\vec{S_{gt}})||_2^2 +
||P(\vec{p^i}) - P_{gt}||_2^2\right)}
\end{align}
Subscript $l_2$ indicates that this loss uses $l_2$ norm for individual vertices. Likewise, we define
\begin{align}\label{eq:loss_l1}
\texttt{Loss}_{\texttt{2d + 3d},~l_1} = \sum_{i}{\left(||\vec{S}(\vec{p^i}) -\vec{S_{gt}})||_1 + ||P(\vec{p^i}) - 
P_{gt}||_1\right)}
\end{align}
We provide details of $\vec{S}(\vec{p^i})$ construction and $P(\vec{p^i})$ projection in the next subsection.
%
%
	\subsection{Morphable Model}\label{sec:MM}
%
%
\textbf{Geometry Model.} Facial geometries are represented as meshes. Morphable models allow to generate variability 
in both face identity and expression. This is done by adding parametrized displacements to a template face model called 
the mean shape. We use the mean shape and $80$ modes from Basel Face Model \cite{BaselMM} to  generate identities and 
$29$ modes obtained from Face Warehouse dataset \cite{FW} to generate expressions. The meshes are controlled by two 
parameter vectors $\vec{\alpha}_{id} \in \mathbb{R}^{80}$  and $\vec{\alpha}_{exp} \in \mathbb{R}^{29}$:
\begin{align}\label{eq:MM}
  \vec{S} = \vec{M} + A_{id} \cdot \vec{\alpha}_{id} + A_{exp} \cdot \vec{\alpha}_{exp}.
\end{align}
Vector $\vec{S} \in \mathbb{R}^{3\cdot N}$ stores the coordinates of $N$ mesh vertices. $\vec{M}$ is the mean shape. 
Matrices $A_{id} \in \mathbb{R}^{3\cdot N \times 80}$, $A_{exp} \in \mathbb{R}^{3\cdot N 
\times 29}$ are the modes of variation.
\\
\\
\textbf{Projection Model.} Projection model translates face mesh from the 3d space to a 2d plane. Rotation matrix $R$ 
and translation vector $\vec{t}$ apply a rigid transformation to the mesh. Projection matrix with three parameters $f$, 
$Px$, $Py$ transforms mesh coordinates to the homogeneous space. For a vertex $\vec{v} = \left(x_m, y_m, z_m\right)^T$ the 
transformation is defined as:
\begin{align}\label{eq:Rt}
\left(\begin{array}{c} x_t \\ 
                        y_t \\ 
                        z_t \end{array}\right) =
\Pi \cdot
\left(\begin{array}{cc} R & ~\vec{t} \end{array}\right) \cdot
\left(\begin{array}{c} x_m \\
                       y_m \\
                       z_m \\
                       1\end{array}\right),                                               
~~~\Pi = \left(\begin{array}{ccc} f & ~0 & ~P_x \\ 0 & ~f & ~P_y \\ 0 & ~0 & ~1 \end{array}\right),
\end{align}
and the final projection of a vertex to the image plane is defined by $u$ and $v$ as:
\begin{align}\label{eq:uv}
u = \frac{x_t}{z_t},~~~v = \frac{y_t}{z_t}.
\end{align}
The projection is defined by $9$ parameters including three rotation angles, three translations and three 
parameters of the projection matrix $\Pi$. We denote projected coordinates by:
\begin{align}\label{eq:P}
P(\Pi, R, \vec{t}, \vec{S}) = \left(\begin{array}{cccc} u_1 & u_2 & \ldots & u_N \\ v_1 & v_2 & \ldots & v_N \end{array}
\right)^T
\end{align}
%
%
	\subsection{Data Preparation}
%
%
Our objective here is to produce a dataset of image-model pairs for neural network training. We use the 
fitting algorithm detailed in Sec. 
\ref{sec:opt} to process the 300W database of annotated face images \cite{300W}. Despite its accuracy reported in Sec. 
\ref{sec:exp_fitting} this algorithm has two limitations. First, the expressive power of the morphable model is 
inherently limited due to laboratory conditions in which the model was obtained and due to the lighting model being 
used. Hence, the model can't generate occlusions and extreme lighting conditions. Second, the hyperparameters of the 
algorithm have been tuned for a dataset taken under controlled conditions. Due to these limitations, the algorithm 
inevitably fails on some of the in-the-wild photos. To overcome this shortcoming, we visually inspect the results and 
delete failed photos. Note that we do not use any specific criteria and this deletion is guided by the visual appeal 
of the models, hence it may be performed by an untrained individual. This leaves us with an even smaller amount of 
images than has initially been in the 300W dataset, namely $2300$ images.
\renewcommand{\arraystretch}{0}
\begin{figure}
\begin{center}
\begin{tabular}{ccccc}
\subfloat{\includegraphics[width = \figsz\textwidth]{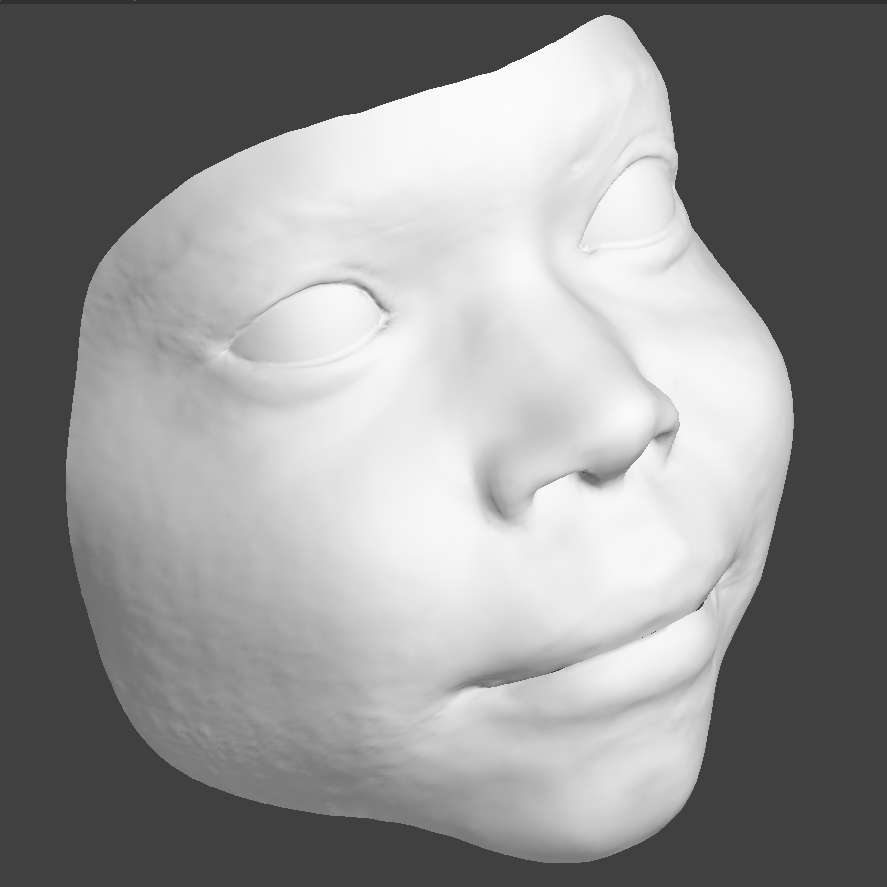}} &
\subfloat{\includegraphics[width = \figsz\textwidth]{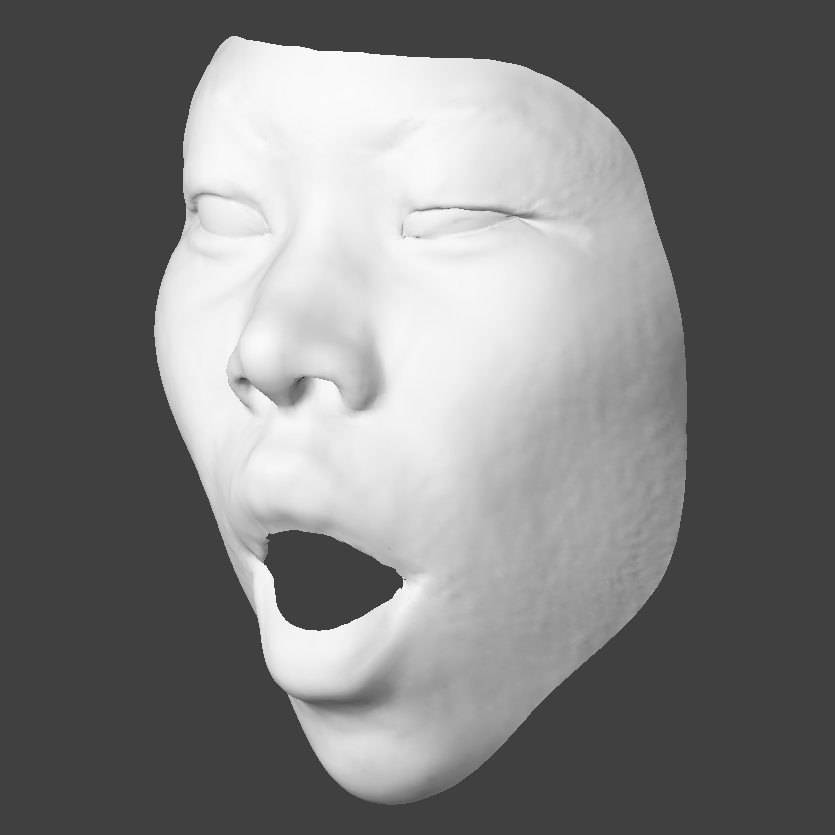}} &
\subfloat{\includegraphics[width = \figsz\textwidth]{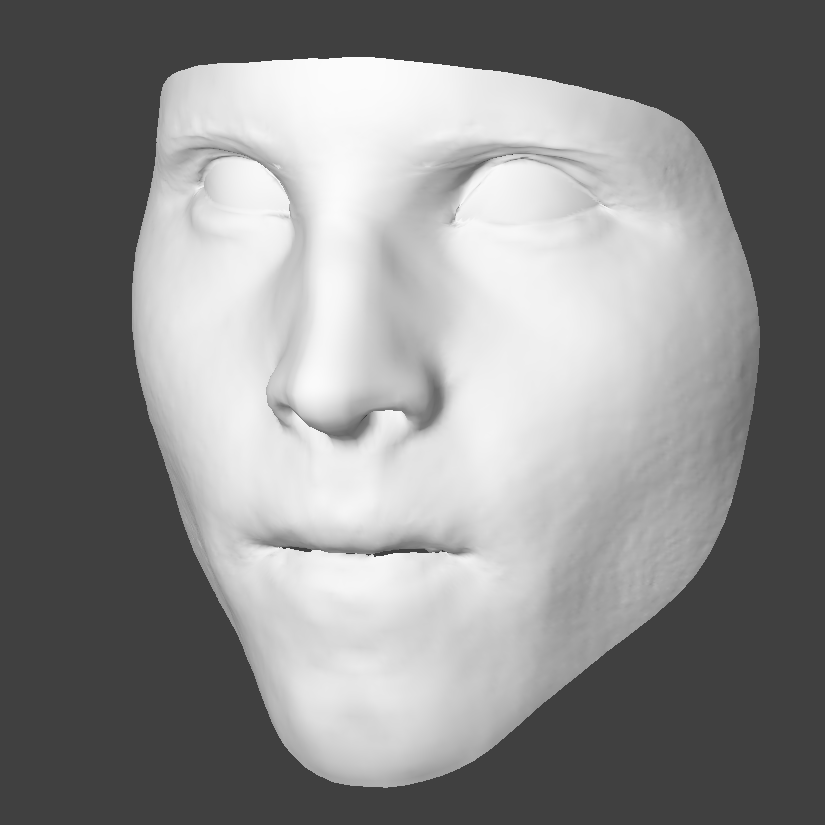}} &
\subfloat{\includegraphics[width = \figsz\textwidth]{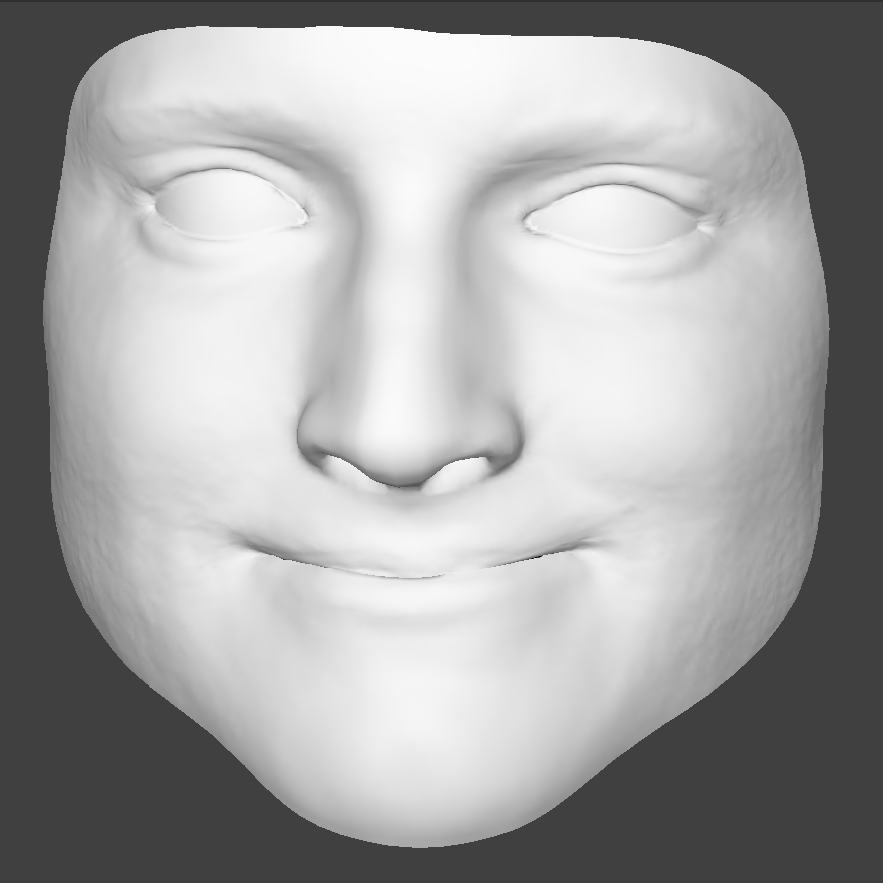}} &
\subfloat{\includegraphics[width = \figsz\textwidth]{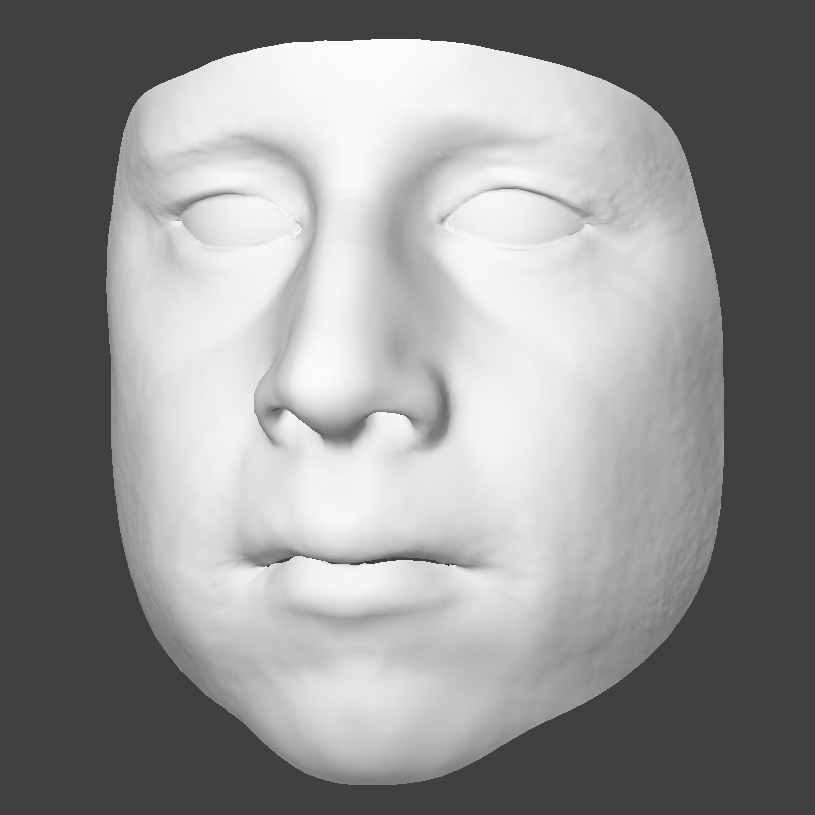}}\\
\subfloat{\includegraphics[width = \figsz\textwidth]{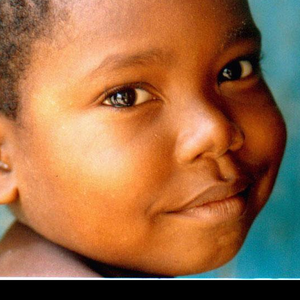}} &
\subfloat{\includegraphics[width = \figsz\textwidth]{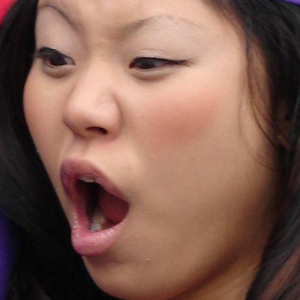}} &
\subfloat{\includegraphics[width = \figsz\textwidth]{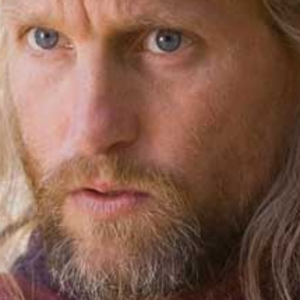}} &
\subfloat{\includegraphics[width = \figsz\textwidth]{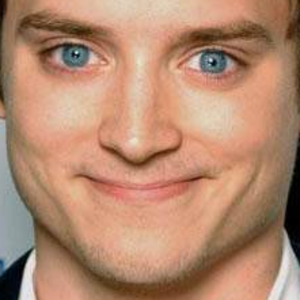}} &
\subfloat{\includegraphics[width = \figsz\textwidth]{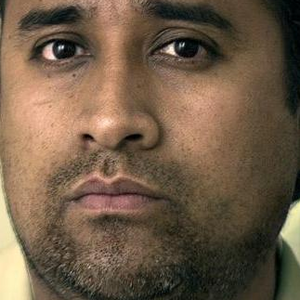}}\\
\end{tabular}
\caption{Example images and corresponding curated ground truth from our training set.}
\label{fig:exTrain}
\end{center}
\end{figure}
\renewcommand{\arraystretch}{1}
This necessitates data augmentation. We 
randomly add blur and noise in both RGB and HSV spaces. Since some of the images with large occlusions have been 
deleted during visual inspection, we compensate for this and randomly occlude images with black rectangles of varied 
sizes~\cite{cutout}. Fig.~\ref{fig:exTrain} shows some examples of our training images.
%
%
	\subsection{Network Architecture}
%
%
Architecture of our network is based on MobileNet \cite{MobileNet}. It consists of interleaving convolution and 
depth-wise convolution \cite{DepthWise} layers followed by average pooling and one fully connected layer. Each 
convolution layer is followed by a batch normalization step \cite{BatchNorm} and a ReLU activation. Input images are 
resized to $96 \times 96$. The final fully-connected layer generates the outputs vector $\vec{p}$ eq. (\ref{eq:vectorP}). Main changes 
compared to the original architecture in \cite{MobileNet} include the decreased input image size $96 \times 96 \times 3$, the first convolution 
filter is resized to $3 \times 3 \times 3 \times 10$, the following filters are scaled accordingly, global average pooling is performed 
over $2 \times 2$ region, and the shape of the FC layer is $320 \times 118$.
%
%
\section{Morphable Model Fitting}
%
%
We use morphable model fitting to generate 3d models of real-world faces to be used for neural network training. Our 
implementation follows standard practices \cite{thies2016face, GarridoRigs}. Geometry and projection models have been 
defined in (Sec. \ref{sec:MM}). Texture model and lighting allow to generate face images. Morphable model fitting aims 
to revert the process of image formation by finding the combination of parameters that will result in a synthetic image 
resembling the target image as closely as possible.
%
%
	\subsection{Image Formation}
%
%
\textbf{Texture Model.} Face texture is modeled similarly to eq. (\ref{eq:MM}). Each vertex of the mesh is assigned 
three RGB 
values generated from a linear model controlled by a parameter vector $\vec{\beta}$:
\begin{align}\label{eq:MT}
  \vec{T} = \vec{T}_0 + B \cdot \vec{\beta}.
\end{align}
We use texture mean and modes from BFM \cite{BaselMM}.
\\
\\
\textbf{Lighting Model.} We use the Spherical Harmonics basis \cite{invrend, envmap} for light computation. The 
illumination of a vertex having albedo $\rho$ and normal $\vec{n}$ is computed as 
\begin{align}\label{light}
I = \rho \cdot \left({\begin{array}{cc} \vec{n}^T & 1\end{array}}\right) \cdot M \cdot \left({\begin{array}{c} \vec{n} 
\\ 1\end{array}}\right),
\end{align}
$M$ is as in \cite{envmap} having $9$ controllable parameters per channel. RGB intensities are
computed separately thus giving overall $9 \cdot 3 = 27$ lighting parameters, $\vec{l} \in \mathbb{R}^{27}$ is the 
parameter vector. Albedo $\rho$ is dependent on $\vec{\beta}$ and computed as in eq. (\ref{eq:MT}).
%
%
	\subsection{Energy Function}
%
%
Energy function expresses the discrepancy between the original attributes of an image and the ones generated from the 
morphable model:
\begin{align}\label{texture_term}
\begin{split}
E = E_{\texttt{tex}} + c_{\texttt{lands}} \cdot E_{\texttt{lands}} + r_{beta, 2} \cdot E_{\texttt{reg,tex}}
+ r_{exp, 2} \cdot E_{\texttt{reg,exp}}.
\end{split}
\end{align}
We describe individual terms of this energy function below.
\\
\\
\textbf{Texture.} The texture term $E_{\texttt{tex}}$ measures the difference between the target image and the one rendered from the
model. We translate both rendered and target images to a standardized UV frame as in \cite{BaselFitting} to unify all 
the image resolutions. Visibility mask 
$\mathscr{M}$ cancels out the invisible pixels.
\begin{align}\label{eq:texture}
E_{\texttt{tex}} = \frac{||\mathscr{M} \cdot (I_{\texttt{target}} - I_{\texttt{rendered}})||}{|\mathscr{M}|}.
\end{align}
We produce $I_{\texttt{rendered}}$ by applying eq. (\ref{light}) and $I_{\texttt{target}}$ by sampling from the
target image at the positions of projected vertices $P$ eq. (\ref{eq:P}). Visibility mask $\mathscr{M}$ is computed 
based on the orientations of vertex normals. We test three alternative norms in place of $||\cdot||$: $l_1$, $l_2$ and 
$l_{2,1}$ norm \cite{thies2016face} that sums $l2$ norms computed for individual pixels.
\\
\\
\textbf{Landmarks.} We use the landmark detector of \cite{Kazemi}. Row indices $\mathscr{L} = \{k_i\}_{i = 1}^{68}$ for 
matrix $P$ eq. (\ref{eq:P}) correspond to the $68$ landmarks. Detected landmarks are $L \in \mathbb{R}^{68\times2}$. 
The landmark term is defined as:
\begin{align}\label{eq:lands}
E_{\texttt{lands}} = ||L - P_{\mathscr{L}, :}||_2^2.
\end{align}
One problem with this term is that indices $\mathscr{L}$ are view-dependent due to the landmark marching. We adopt a 
solution similar to that of \cite{HighFidelityNormalization} and annotate parallel lines of vertices for the landmarks 
on the border.
\\
\\
\textbf{Regularization.} We assume multivariate Gaussian priors on morphable model parameters as defined below and use
$\sigma_{id}$ and $\sigma_{tex}$ provided by~\cite{BaselMM}.
\begin{align}\label{reg}
E_{\texttt{reg,id}} = \sum_{i = 1}^{80}{\frac{\alpha_{id, i}^2}{\sigma_{id, i}^2}},
~~ E_{\texttt{reg,exp}} = \sum_{i = 1}^{29}{\frac{\alpha_{exp, i}^2}{\sigma_{exp, i}^2}},
~~ E_{\texttt{reg,tex}} = \sum_{i = 1}^{80}{\frac{\beta_{i}^2}{\sigma_{tex, i}^2}}
\end{align}
We regularize neither lighting nor projection parameters.
%
%
	\subsection{Optimization}\label{sec:opt}
%
%
Optimization process is divided into two major steps: First, we minimize the landmark term:
\begin{align}\label{landmark_term}
E_1 = E_{\texttt{lands}} + r_{id, 1} \cdot E_{\texttt{reg,id}} + r_{exp, 1} \cdot E_{\texttt{reg,exp}}.
\end{align}
We then minimize the full energy function eq. (\ref{texture_term}). These two steps are also divided into sub-steps 
minimizing the energy function with respect to specific parameters similarly to \cite{GarridoRigs}. We minimize the 
energy function with respect to only one type of parameters at any moment.
%
%
%
%
We do not include identity regularization into eq. (\ref{texture_term}) because it did not improve accuracy in 
our experiments.
%
%
\section{Experiments}
%
%
We carry out three sets of experiments. First, we study the effect of different settings for the fitting of the 
morphable model used in this paper. Second, we experiment with different losses and datasets for neural network 
training. Finally, we present a comparison of our method with other recent approaches.

Unfortunately current research in 3d face reconstruction is lacking standardized benchmarks and evaluation protocols.
As a result, evaluations presented in research papers vary in the type of error metrics and datasets 
used (see Table~\ref{table:methodsTestsets}). This makes the results from many works difficult to compare. We hope 
to contribute towards filling this gap by providing the standard evaluation code and a testing set of 
images\footnote{\url{https://github.com/nchinaev/MobileFace}}.

\textbf{BU4DFE Selection.} Tulyakov et al. \cite{tulyakov} provide annotations for a total of $3000$ selected scans 
from BU4DFE. We divide this selection into two equally sized subsets BU4DFE-test and BU4DFE-val. We report final 
results on the former and experiment with hyperparameters on the latter. For the purpose of evaluation we use 
annotations to initialize the ICP alignment.
%
%
	\subsection{Implementation Details}
%
%
We trained networks for the total of $3\cdot 10^5$ iterations with the batches of size $128$. We added $l_2$ weight
decay with coefficient of $10^{-4}$ for regularization. We used Adam optimizer \cite{Adam} with learning rate of 
$10^{-4}$ for iterations before $2\cdot10^5$-th and $10^{-5}$ after. Other settings for the optimizer are standard.
Coefficients for morphable model fitting are $r_{id, 1} = 0.001$, $r_{exp, 1} = 0.1$, $r_{beta, 2} = 0.001$, 
$c_{\texttt{lands}} = 10$, $r_{exp, 2} = 10$.
%
%
	\subsection{Accuracy Evaluation}
%
%
Accuracy of 3D reconstruction is estimated by comparing the resulting 3D model to the ground truth facial scan. To 
compare the models, we first perform ICP alignment. Having reconstructed facial mesh $\vec{S}$ and the 
ground truth scan $\vec{S_{gt}}$, we project vertices of $\vec{S}$ on $\vec{S_{gt}}$ and Procrustes-align $\vec{S}$ to 
the projections. These two steps are iterated until convergence.
\\
\\
\textbf{Error Measure.} To account for variations in scan sizes, we use a normalization term
\begin{align}\label{eq:1}
  C(\vec{S_{gt}}) = ||\vec{S_{gt}^0}||_2^2,
\end{align}
where $\vec{S_{gt}^0}$ is $\vec{S_{gt}}$ with the mean of each x, y, z coordinate subtracted. The dissimilarity measure
between $\vec{S}$ and $\vec{S_{gt}}$ is
\begin{align}\label{eq:errorMetric}
  d(\vec{S}, \vec{S_{gt}}) = c_s \cdot\frac{||\vec{S} - \vec{S_{gt}}||_2^2}{C(\vec{S_{gt}})}
\end{align}
The scaling factor $c_s = 100$ is included for convenience.
\setlength{\tabcolsep}{4pt}
\begin{table}
\begin{center}
\caption{Methods and their corresponding testsets.}
\label{table:methodsTestsets}
\begin{tabular}{c|c}
\hline\noalign{\smallskip}
Work & Testset\\
\noalign{\smallskip}
\hline
\noalign{\smallskip}
Jackson et al. \cite{VRN} & AFLW2000-3D, renders from BU4DFE and MICC\\
Tran et al. \cite{tran2017regressing} & MICC video frames \\
Tewari et al. \cite{MOFA} & synthetic data; Face Warehouse \\
Dou et al. \cite{dou2017end} & UHDB31, FRGC2, BU-3DFE \\
Roth et al. \cite{roth2016adaptive} & renders from BU4DFE \\
 & \\
\end{tabular}
\end{center}
\end{table}
\setlength{\tabcolsep}{1.4pt}
%
%
%
%
%
	\subsection{Morphable Model Fitting}\label{sec:exp_fitting}
\begin{figure}
	\centering
	\includegraphics[height=6.5cm]{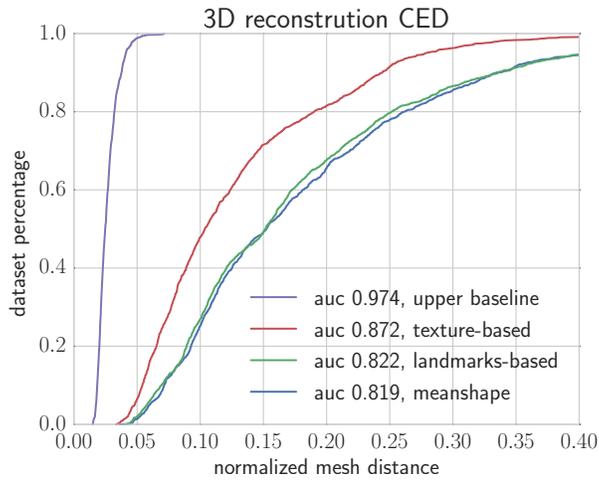}
	\caption{Evaluation of fitting methods on BU4DFE-test. Areas under curve are computed for normalized mesh 
			 distances ranging from $0$ to $1$. Shorter span of x-axis is used for visual clarity.}
	\label{fig:landsVStex}
\end{figure}

\begin{figure}
	\centering
	\subfloat[Evaluations for different norms. $r_{exp, 2} = 0$] 
             {\includegraphics[width=0.48\textwidth]{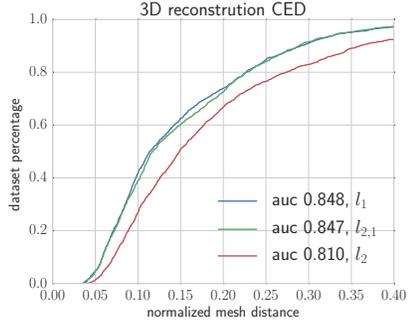}\label{fig:texVariants1}}
	\hfill
	\subfloat[Evaluations with different regularizations.
	          $l_{2, 1}$ norm is used in all cases.] 
             {\includegraphics[width=0.48\textwidth]{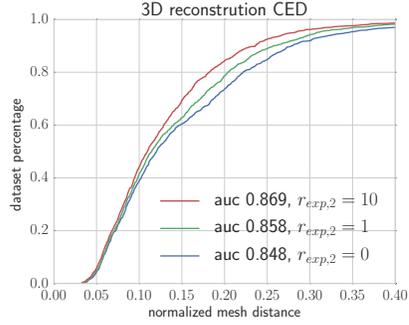}\label{fig:texVariants2}}
	\hfill
	\caption{Evaluation of texture-based fitting algorithm on BU4DFE-val with different settings. }
\end{figure}

%
%
We compare the accuracy of the fitting algorithm in two major settings: using only landmarks and using landmarks in 
combination with texture. To put the numbers in a context, we establish two baselines. First baseline is attained by 
computing the reconstruction error for the mean shape. This demonstrates the performance of a hypothetical 
dummy algorithm that always outputs the mean shape for any input. Second baseline is computed by registering the 
morphable model to the scans in 3d. It demonstrates the performance of a hypothetical best method that is only bounded 
by the descriptive power of the morphable model.
Landmark-based fitting is done by optimizing eq. (\ref{landmark_term}) from sec. \ref{sec:opt}. Texture-based fitting 
is done by optimizing both eq. (\ref{landmark_term}) and eq. (\ref{texture_term}). Fig. \ref{fig:landsVStex} shows 
cumulative error distributions. It is clear from the graph that texture-based fitting significantly outperforms 
landmark-based fitting which is only as accurate as the meanshape baseline. However, there is still a wide gap between 
the performance of the texture-based fitting and the theoretical limit.
Figs. \ref{fig:texVariants1}, \ref{fig:texVariants2} show the performance of texture-based fitting algorithm with 
different settings. The settings differ in the type of norm being used for texture term computation and the amount of 
regularization. In particular, 
Fig. \ref{fig:texVariants1} demonstrates that the  choice of the norm plays an important role with $l_{2, 1}$ and $l_1$ 
norms outperforming $l_2$. Fig. \ref{fig:texVariants2} shows that the algorithm is quite sensitive to the 
regularization, hence the regularization coefficients need to be carefully tuned.
%
%
%
%
	\subsection{Neural Network}
\begin{figure}
	\centering
	\subfloat[Comparison of networks trained on our cleaned dataset with different losses.] 
{\includegraphics[width=0.48\textwidth]{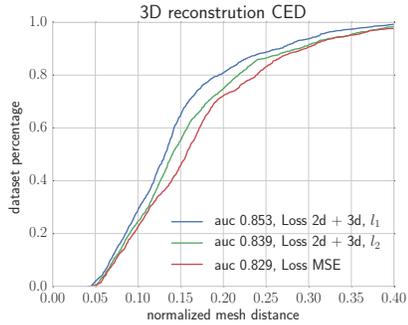}\label{fig:cedExperiments1}}
	\hfill
	\subfloat[Comparison of networks trained with $\texttt{Loss}_{\texttt{2d + 3d},~l_2}$ on different datasets.] 
{\includegraphics[width=0.48\textwidth]{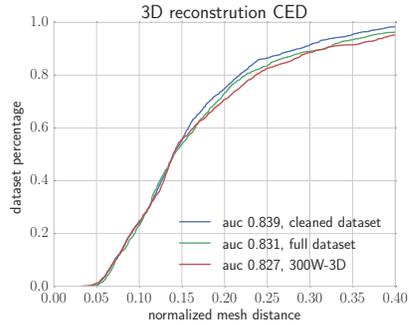}\label{fig:cedExperiments2}}
	\hfill
	\caption{Evaluation on BU4DFE-val for networks trained with different losses on different datasets.}
\end{figure}

%
%
We train the network on our dataset of image-model pairs. For the sake of comparison, we also train it on 300W-3D 
\cite{3DDFA}. The training is performed in different settings: using different loss functions and using manually 
cleaned version of the dataset versus non-cleaned. The tests are performed on BU4DFE-val. Figs. 
\ref{fig:cedExperiments1}, \ref{fig:cedExperiments2} show cumulative error distributions. These experiments support 
following claims:
\begin{itemize}
	\item Learning from our dataset gives better results than learning from 300W-3D,
	\item Our loss function improves results compared to baseline MSE loss function,
	\item Manual deletion of failed photos by an untrained individual improves results.
\end{itemize}
%
%
%
%
%
	\subsection{Comparison with the State of the Art}
%
%
\begin{figure}
	\centering
	\includegraphics[height=6.5cm]{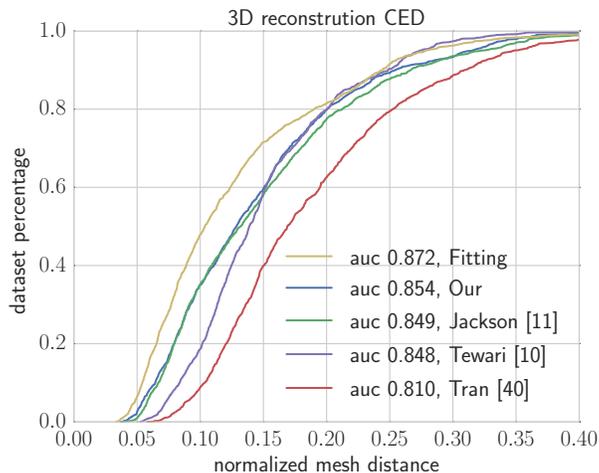}
	\caption{Comparison of methods on BU4DFE-test.}
	\label{fig:cedComparison}
\end{figure}
\noindent
\textbf{Quantitative Results.} Fig. \ref{fig:cedComparison} presents evaluations of our network and a few recent 
methods on BU4DFE-test. Error metric is as in eq. (\ref{eq:errorMetric}). The work of Tran
et al. \cite{tran2017extreme} is based on \cite{tran2017regressing} and their code allows to produce 
non-neutral models, therefore we present an evaluation of \cite{tran2017extreme} and not \cite{tran2017regressing}. We 
do not present an evaluation of 3DDFA \cite{3DDFA} because Jackson et al. \cite{VRN} have already demonstrated that 
3DDFA is inferior to their method. We do not include the work of Sela et al. \cite{Sela} into comparison because we 
were not able to reproduce their results. For Jackson et al. \cite{VRN} we were able to reproduce their error on 
AFLW2000-3D. We used MATLAB implementation of isosurface algorithm to transform their volumes into meshes. Tran et. al 
\cite{tran2017extreme} do not present an evaluation on 3d scans, however we were able to roughly reproduce an error for 
MICC dataset \cite{MICC} from their earlier work \cite{tran2017regressing}. We noticed that their method is sensitive 
to the exact selection of frames from MICC videos. For an optimal selection of frames the error equals $1.43$ which is 
less than $1.57$ reported in \cite{tran2017regressing}. In the worst case error equals $2.34$. Tewari et al. 
\cite{MOFA} did not open-source their implementation of MOFA, but authors kindly provided their reconstructed models 
for our testset.

It is seen from the graph that our network performs on a par with other recent methods being slightly ahead of the 
second-best method. Additionally, the size of our model is orders of magnitude smaller, see Table 
\ref{table:comparisons} for a comparison. We used Intel Core i5-4460 for CPU experiments, NVIDIA GeForce GTX 1080 for 
GPU experiments (except for Tewari \cite{MOFA}, they used NVIDIA Titan X Pascal) and Samsung Galaxy S7 for ARM 
experiments.

Table \ref{table:hassnerMicc} presents a comparison with Tran et al. \cite{tran2017regressing} on MICC dataset
\cite{MICC}. This is a dataset of $53$ subjects. For each of the subjects it provides three videos and a neutral facial 
scan. It is therefore crucial that a method being evaluated on this dataset should output neutral models for any input. 
Method of Tran et al. \cite{tran2017regressing} is specifically designed for this purpose. We adapt our method to this 
scenario by setting $\vec{\alpha}_{exp}$ to zero. We randomly select $23$ frames per individual and form $23$ 
corresponding testsets. We compute errors over these testsets and average those. One important aspect affecting the 
errors is the use of scaling during ICP alignment: Tran et al. \cite{tran2017regressing} did not allow models to scale 
during the alignment. We present evaluations in both settings. 

\begin{figure}
\begin{center}
\begin{tabular}{ccccccc}
Original &
Fitting &
Our &
Jackson\cite{VRN} &
Tewari \cite{MOFA} &
Tran \cite{tran2017extreme} &
GT
\\
\subfloat{\includegraphics[width = \figsz\textwidth]{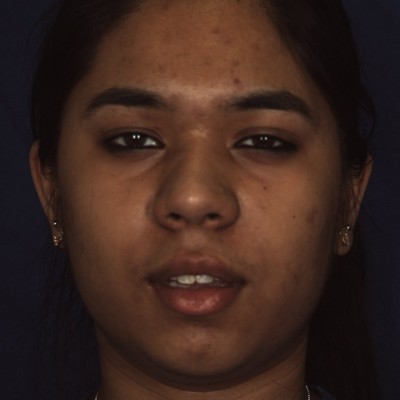}} &
\subfloat{\includegraphics[width = \figsz\textwidth]{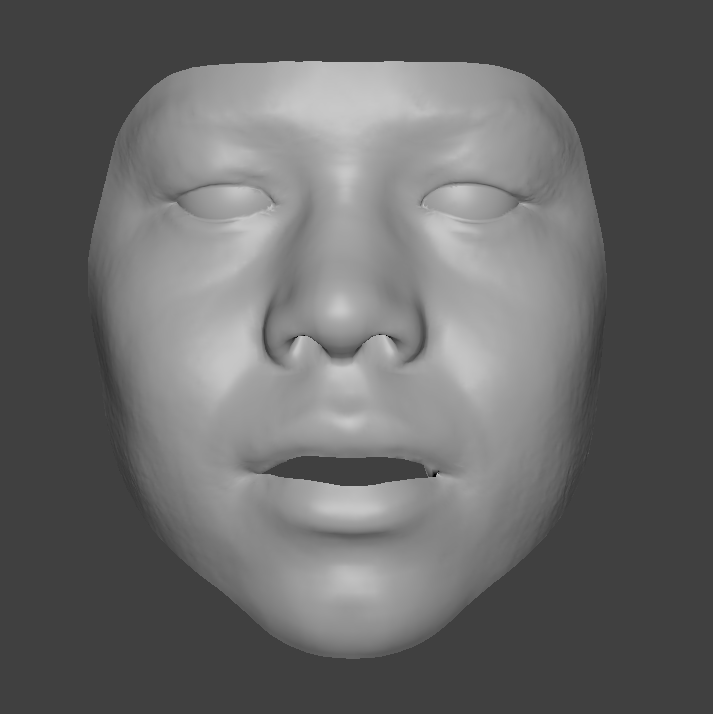}} &
\subfloat{\includegraphics[width = \figsz\textwidth]{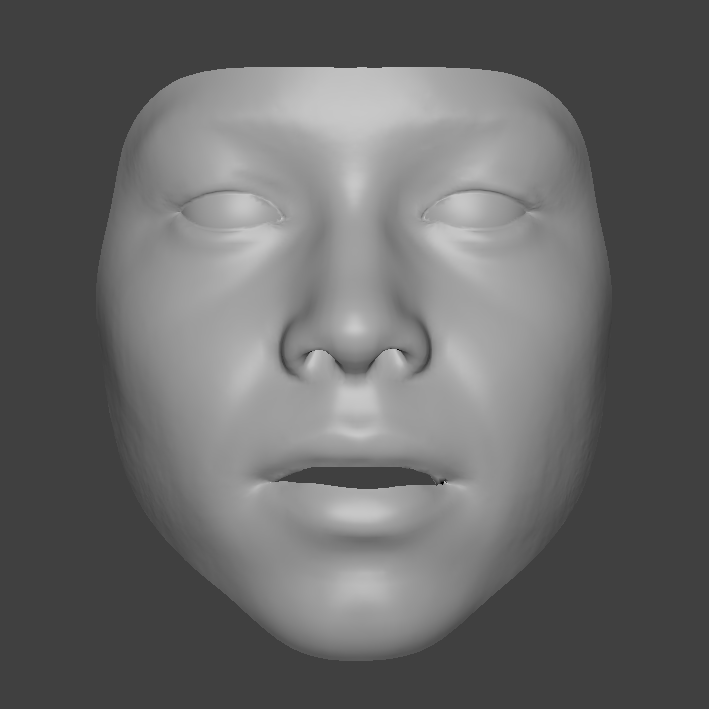}} &
\subfloat{\includegraphics[width = \figsz\textwidth]{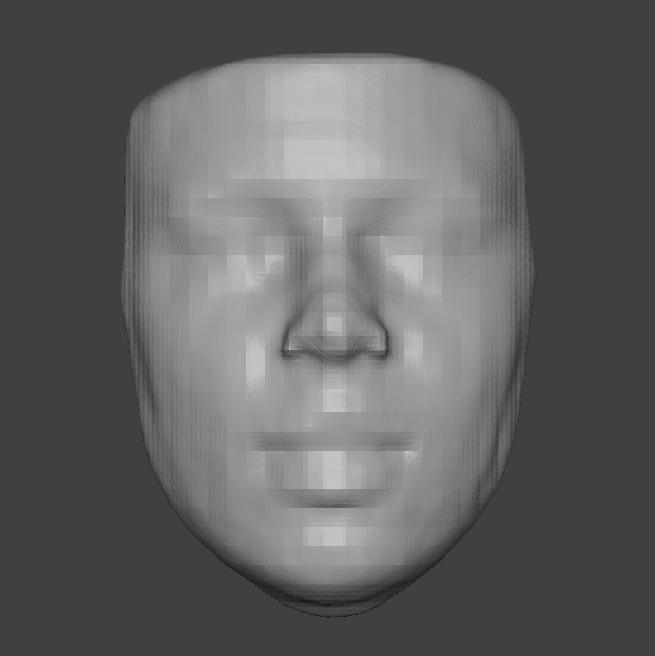}} &
\subfloat{\includegraphics[width = \figsz\textwidth]{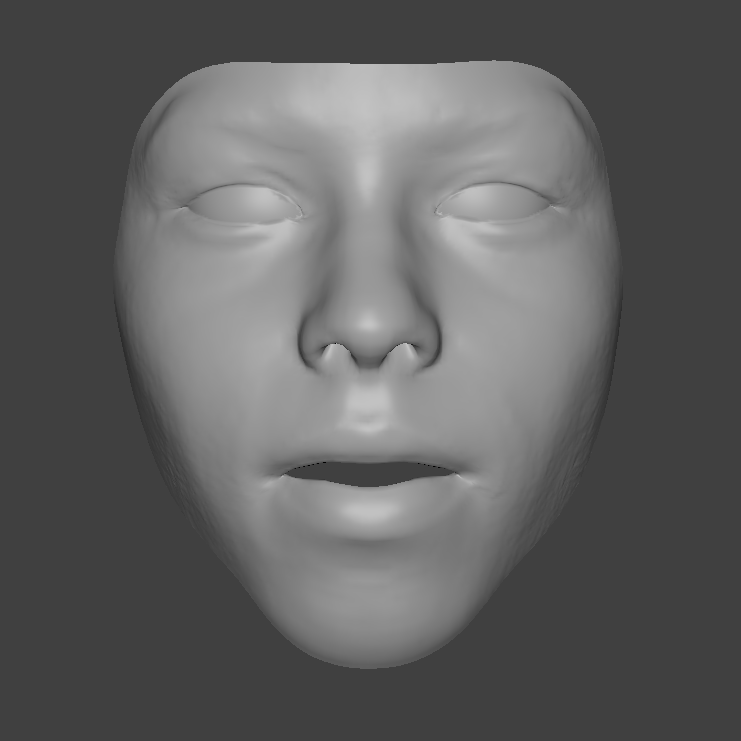}} &
\subfloat{\includegraphics[width = \figsz\textwidth]{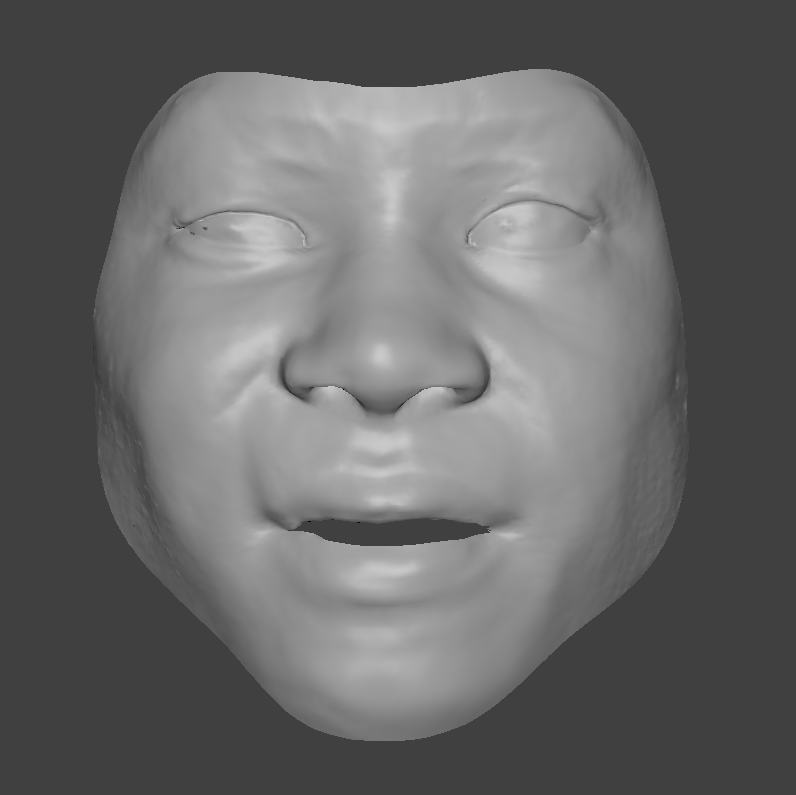}} &
\subfloat{\includegraphics[width = \figsz\textwidth]{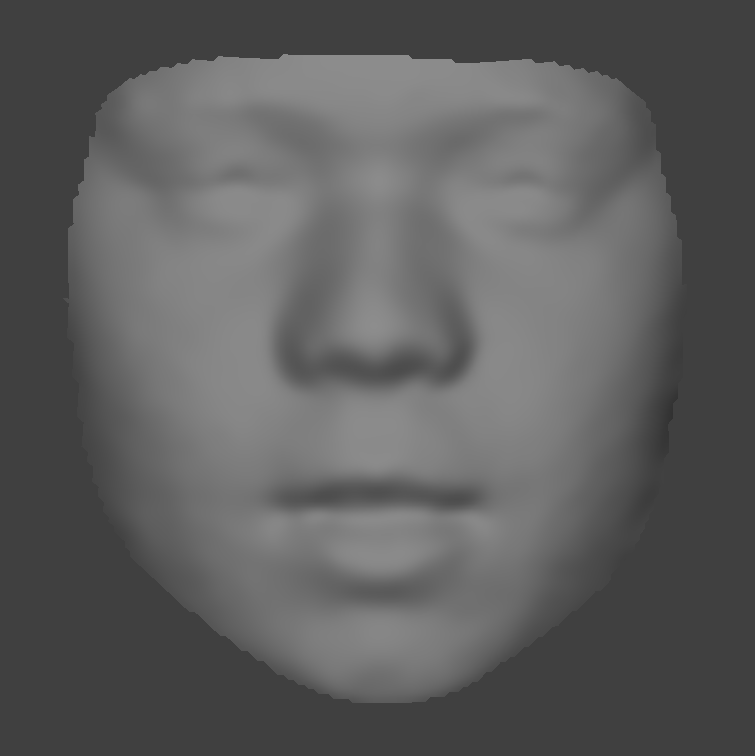}}
\\
\subfloat{\includegraphics[width = \figsz\textwidth]{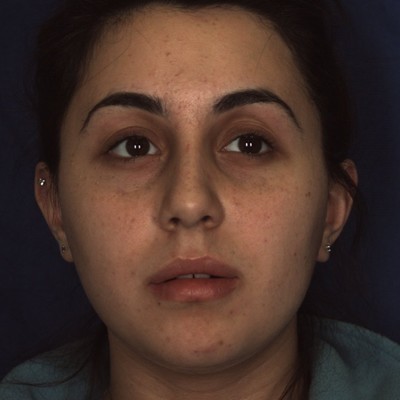}} &
\subfloat{\includegraphics[width = \figsz\textwidth]{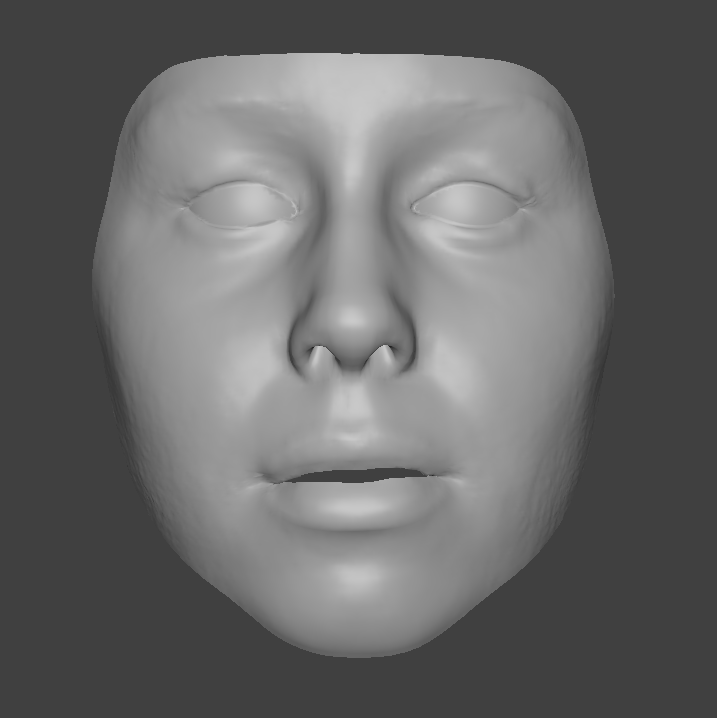}} &
\subfloat{\includegraphics[width = \figsz\textwidth]{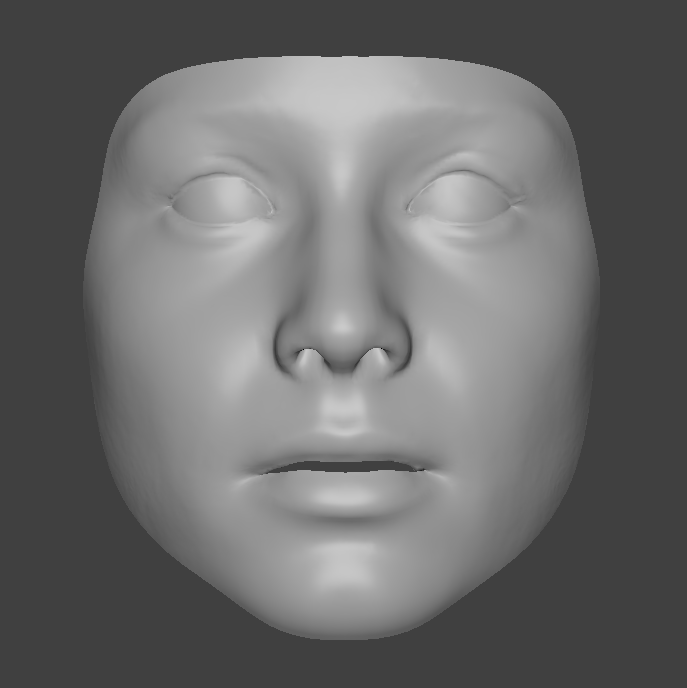}} &
\subfloat{\includegraphics[width = \figsz\textwidth]{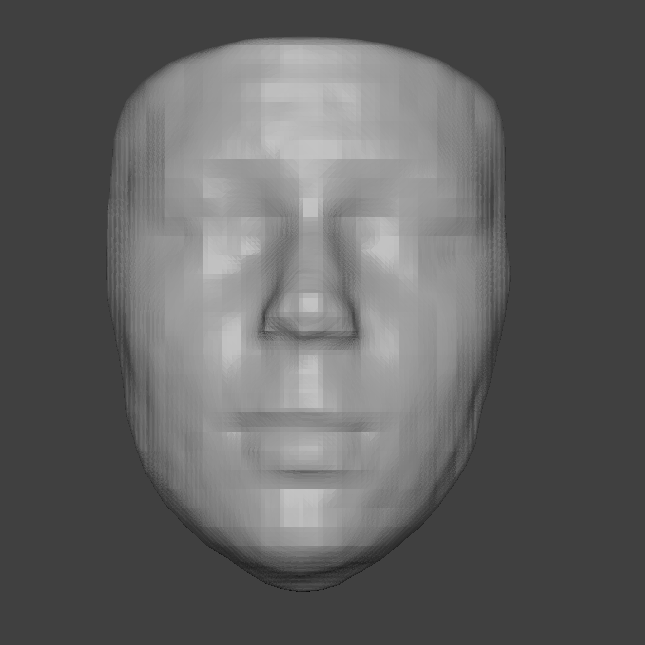}} &
\subfloat{\includegraphics[width = \figsz\textwidth]{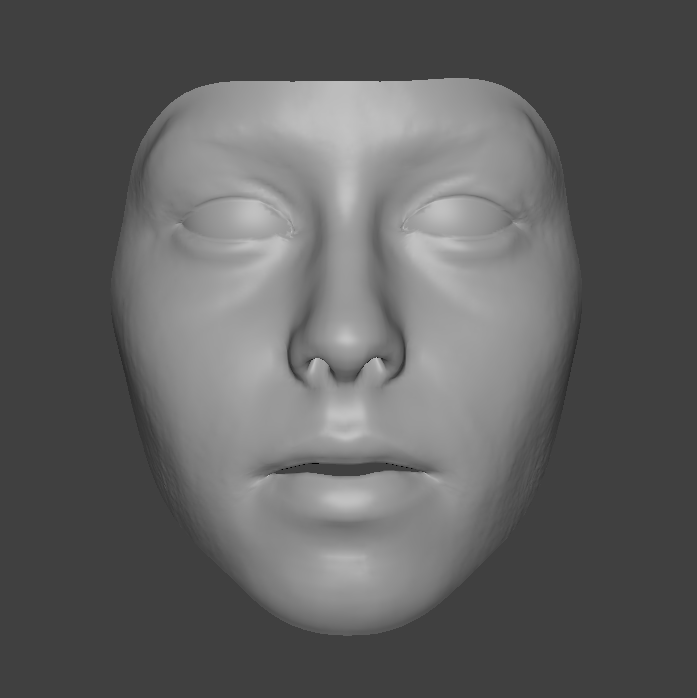}} &
\subfloat{\includegraphics[width = \figsz\textwidth]{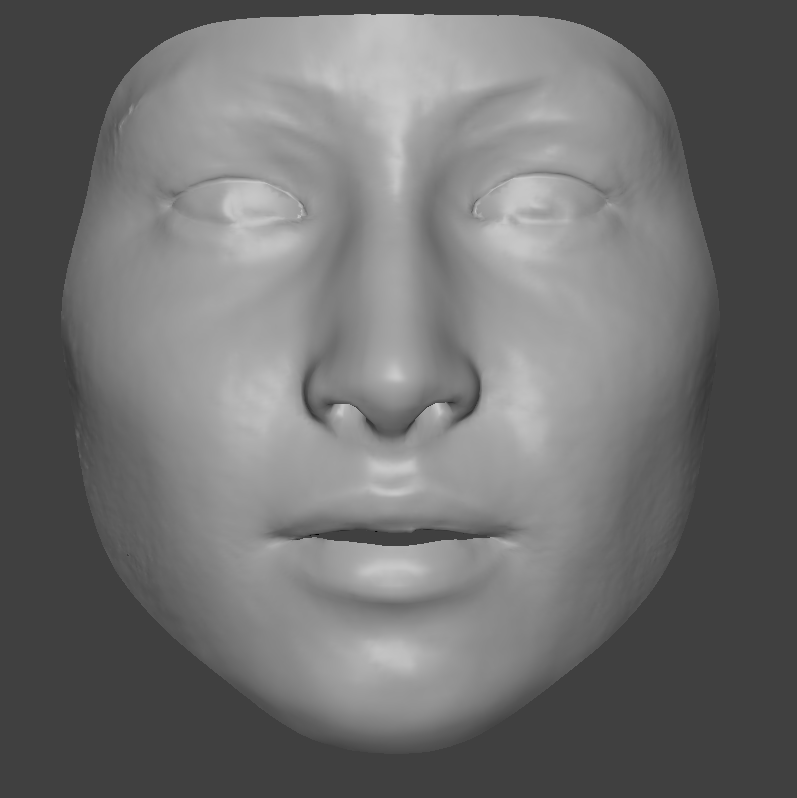}} &
\subfloat{\includegraphics[width = \figsz\textwidth]{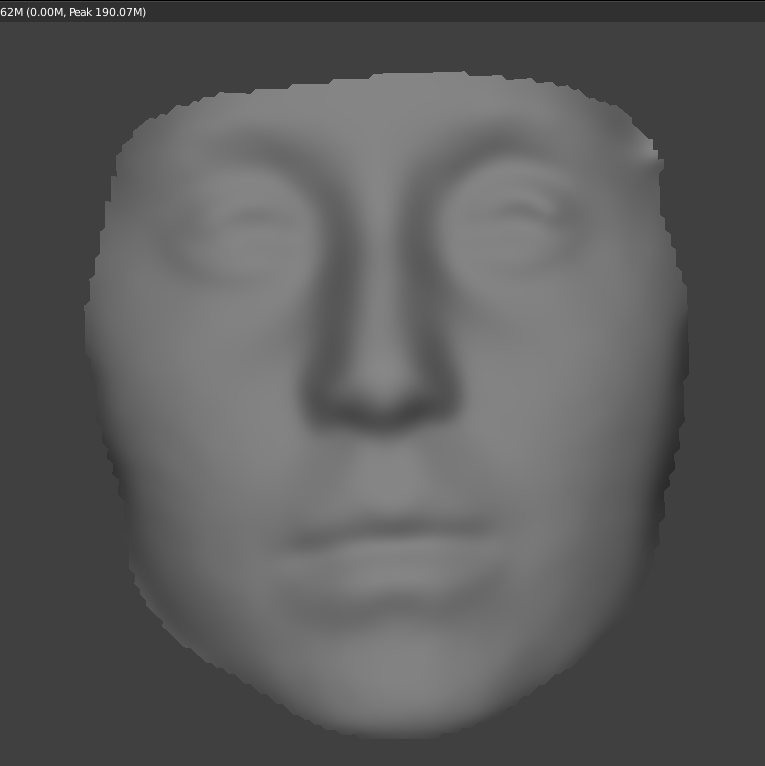}}
\\
\subfloat{\includegraphics[width = \figsz\textwidth]{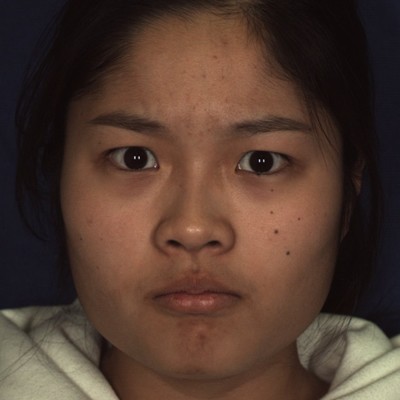}} &
\subfloat{\includegraphics[width = \figsz\textwidth]{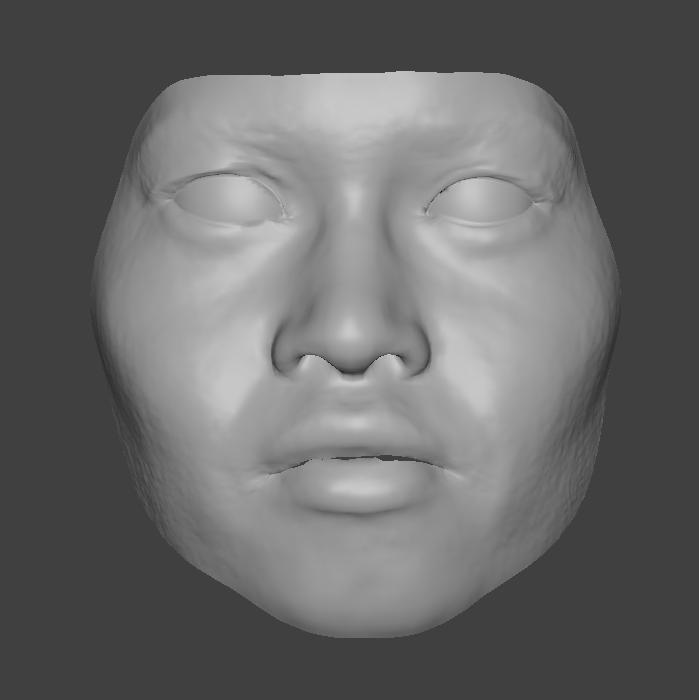}} &
\subfloat{\includegraphics[width = \figsz\textwidth]{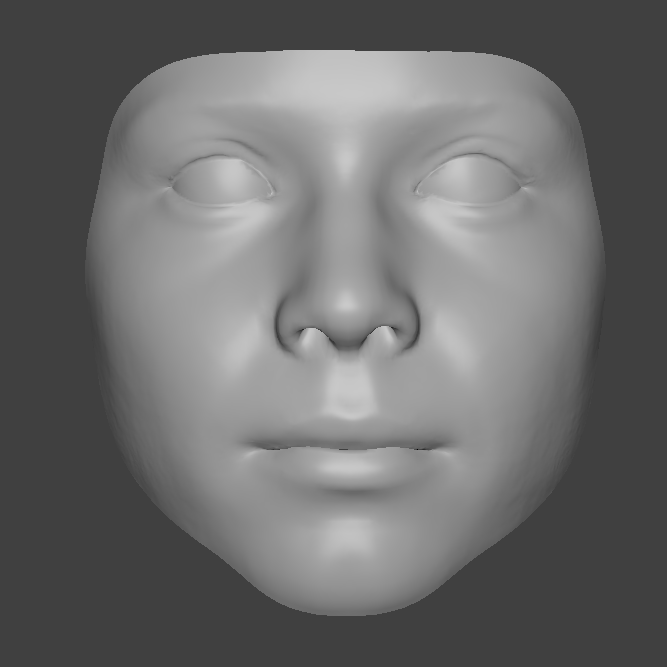}} &
\subfloat{\includegraphics[width = \figsz\textwidth]{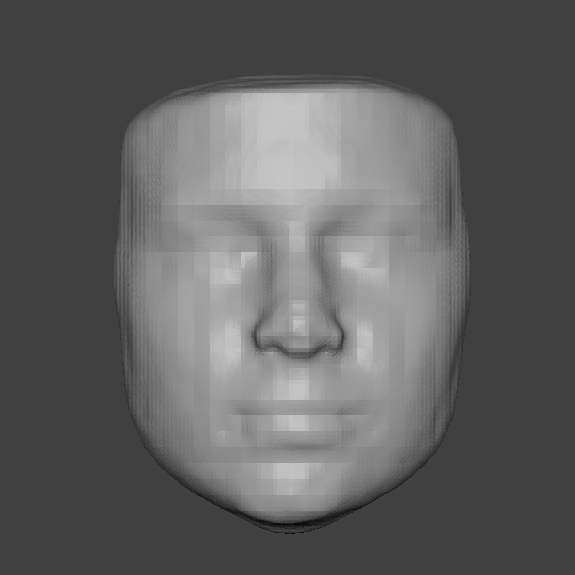}} &
\subfloat{\includegraphics[width = \figsz\textwidth]{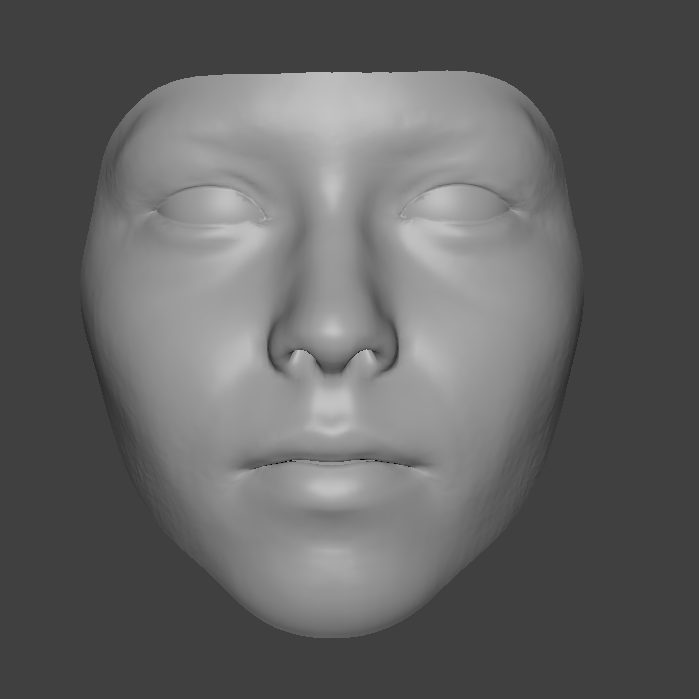}} &
\subfloat{\includegraphics[width = \figsz\textwidth]{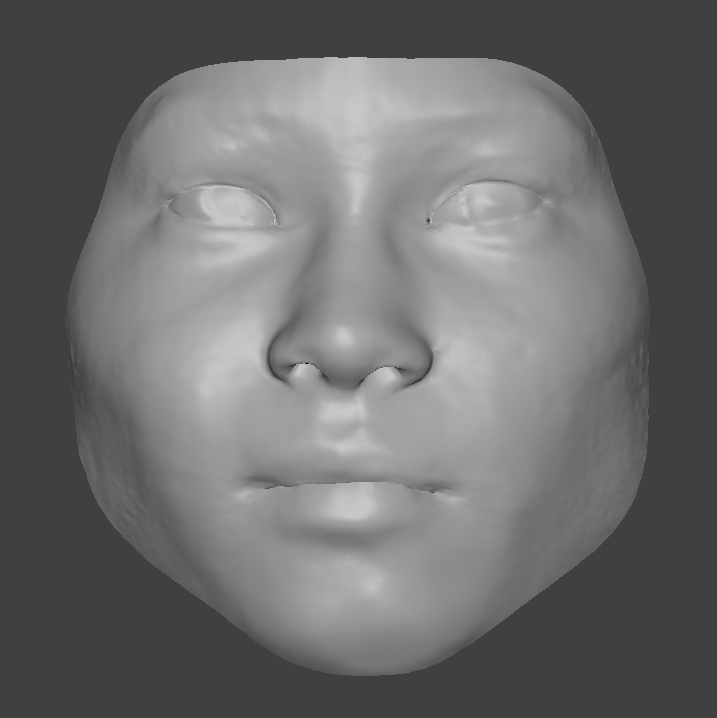}} &
\subfloat{\includegraphics[width = \figsz\textwidth]{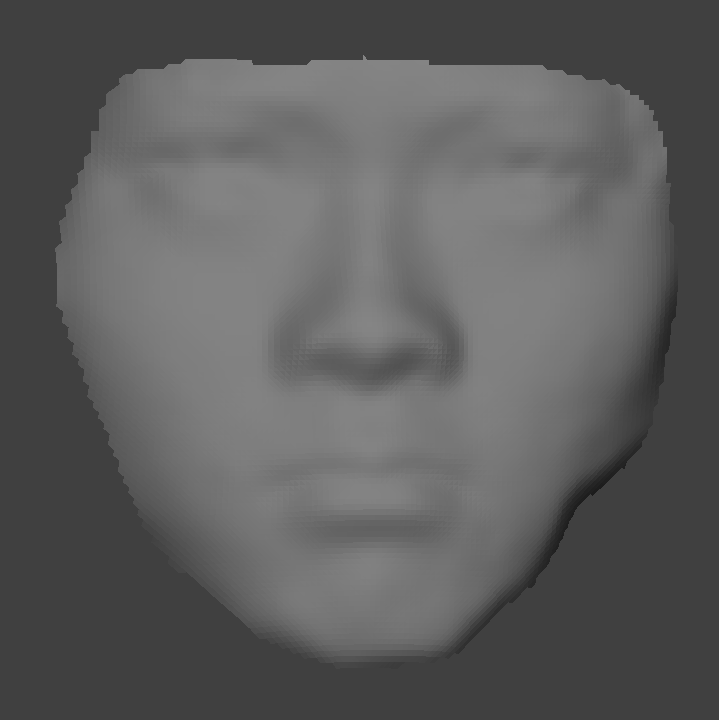}}
\\
\subfloat{\includegraphics[width = \figsz\textwidth]{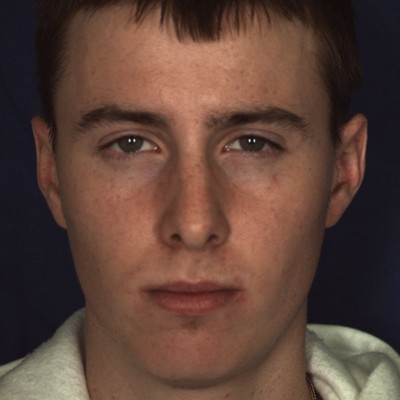}} &
\subfloat{\includegraphics[width = \figsz\textwidth]{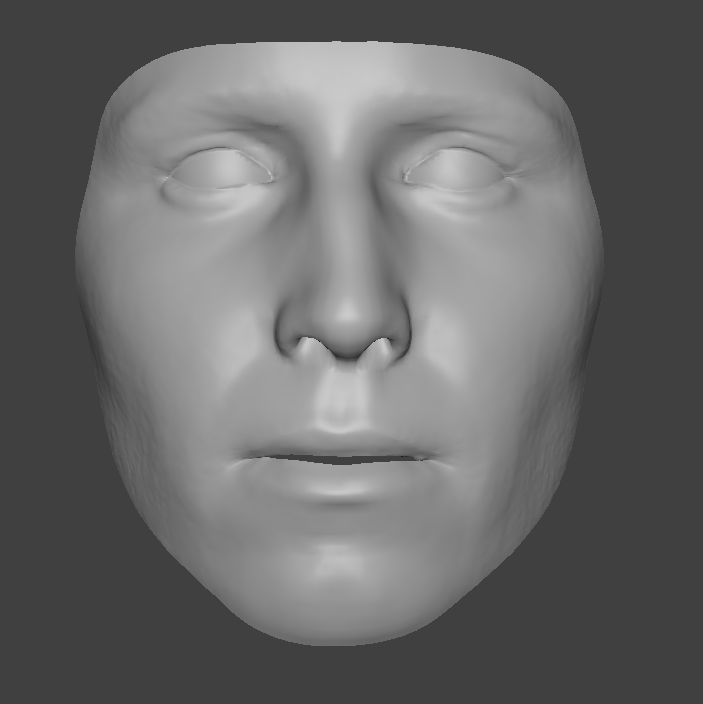}} &
\subfloat{\includegraphics[width = \figsz\textwidth]{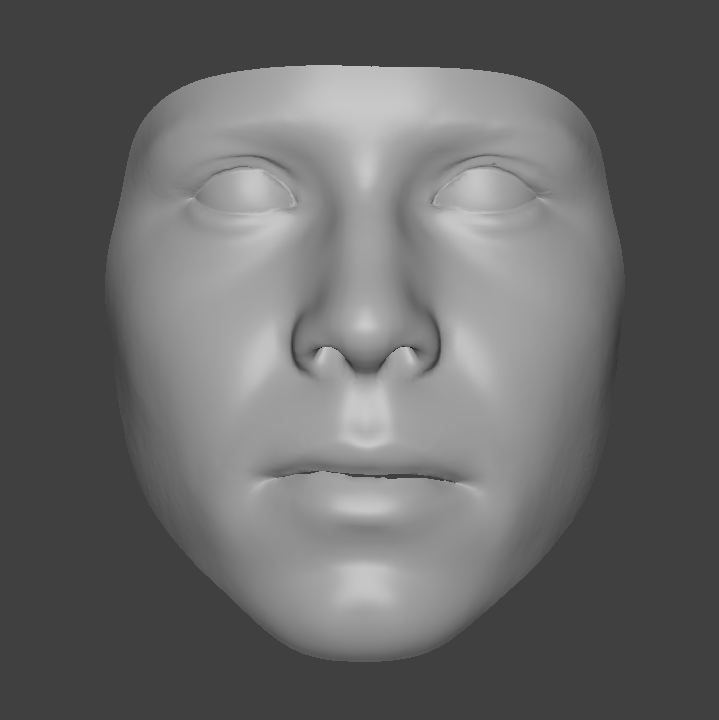}} &
\subfloat{\includegraphics[width = \figsz\textwidth]{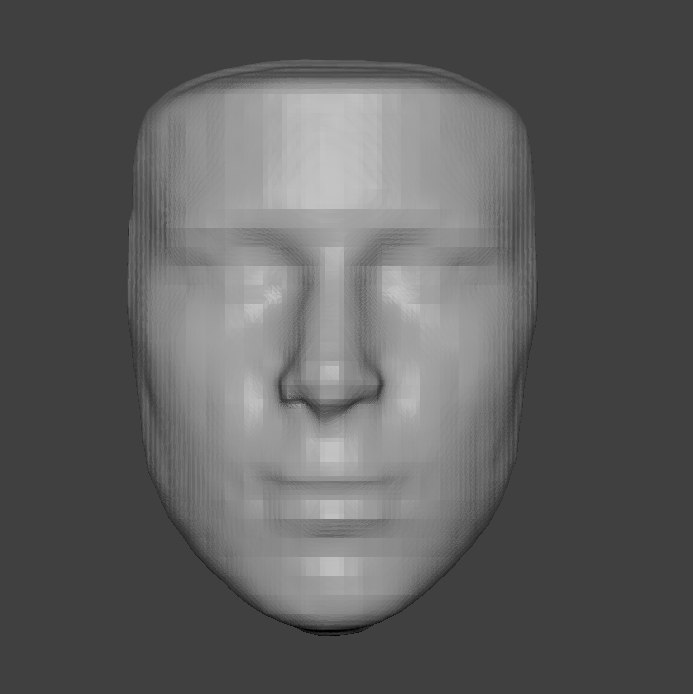}} &
\subfloat{\includegraphics[width = \figsz\textwidth]{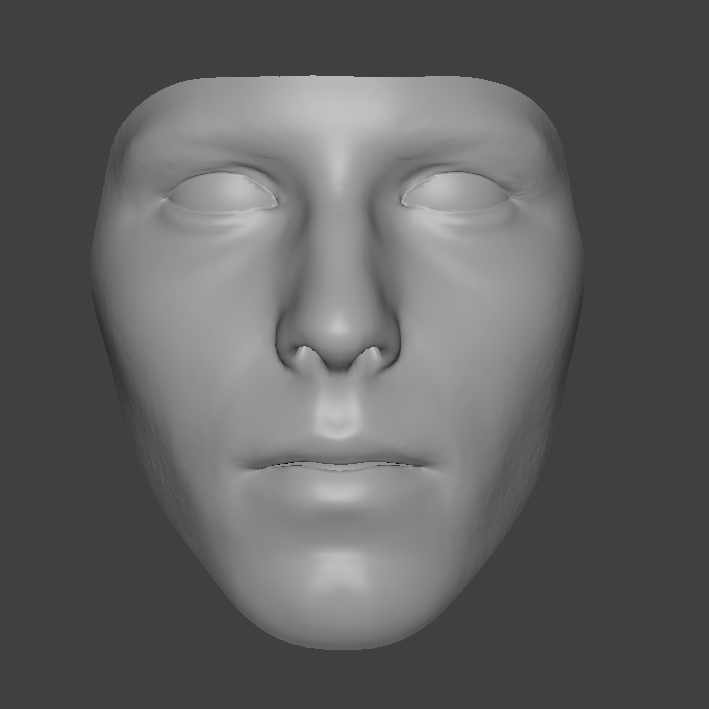}} &
\subfloat{\includegraphics[width = \figsz\textwidth]{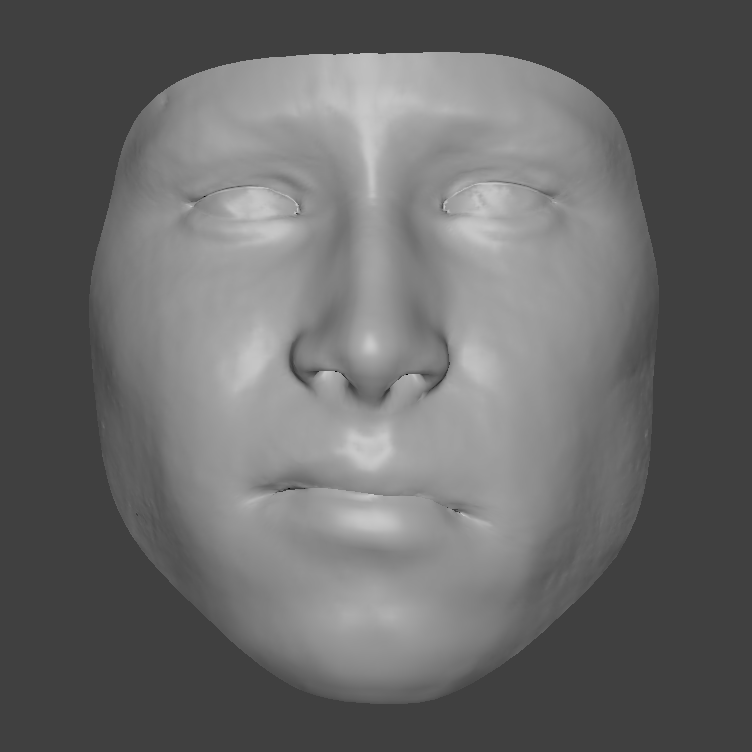}} &
\subfloat{\includegraphics[width = \figsz\textwidth]{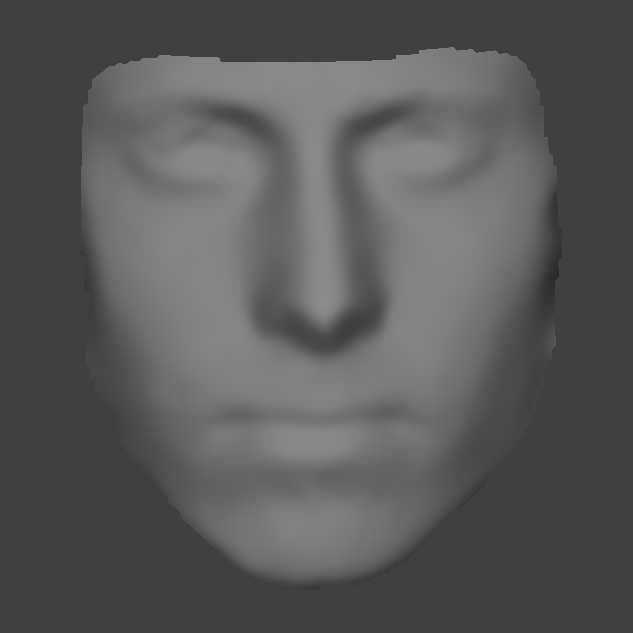}}
\\
\subfloat{\includegraphics[width = \figsz\textwidth]{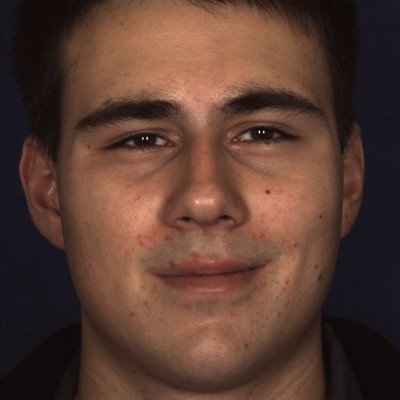}} &
\subfloat{\includegraphics[width = \figsz\textwidth]{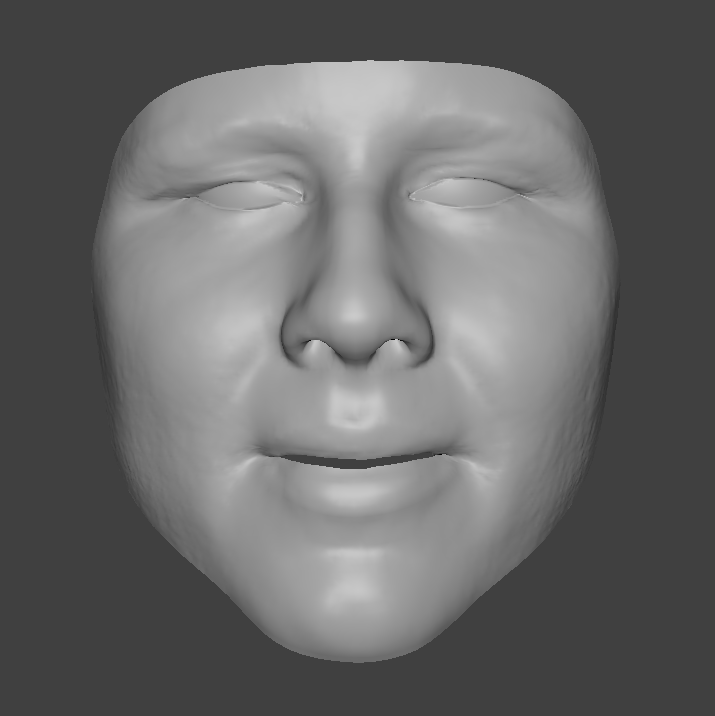}} &
\subfloat{\includegraphics[width = \figsz\textwidth]{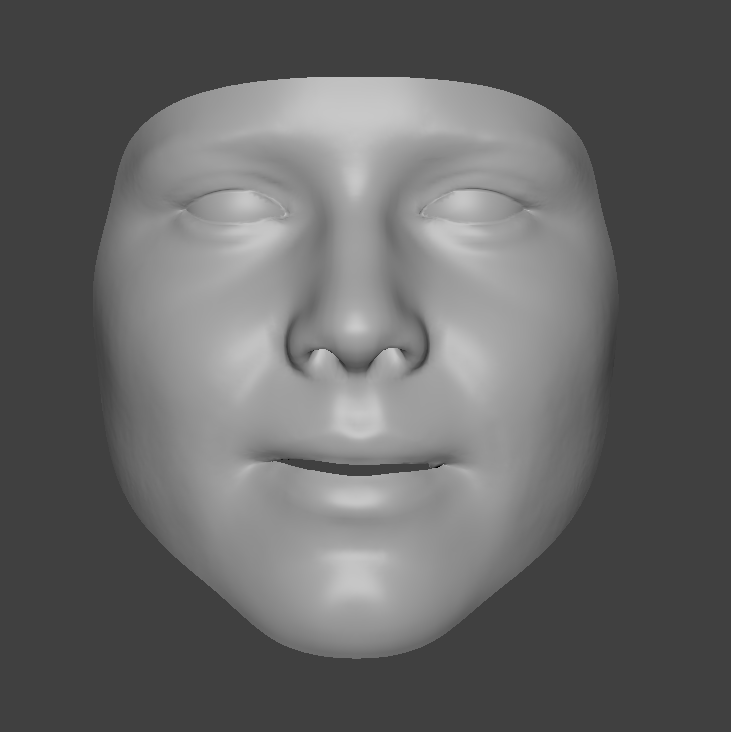}} &
\subfloat{\includegraphics[width = \figsz\textwidth]{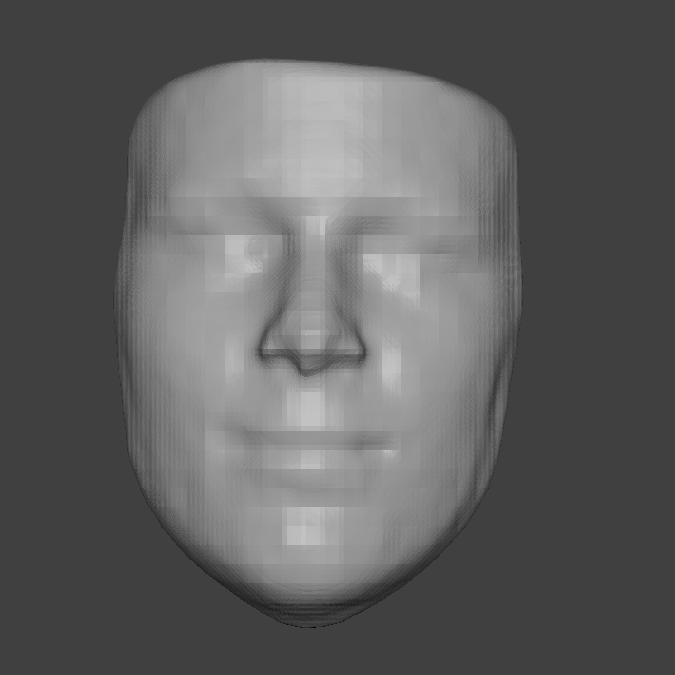}} &
\subfloat{\includegraphics[width = \figsz\textwidth]{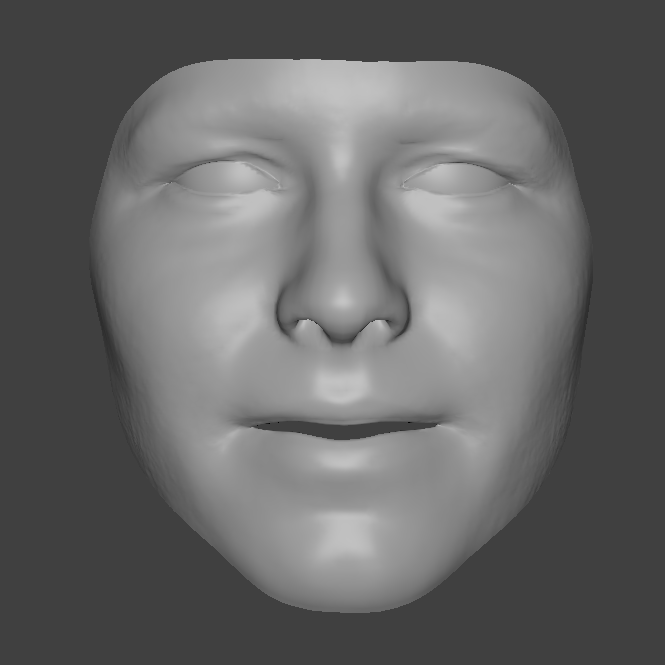}} &
\subfloat{\includegraphics[width = \figsz\textwidth]{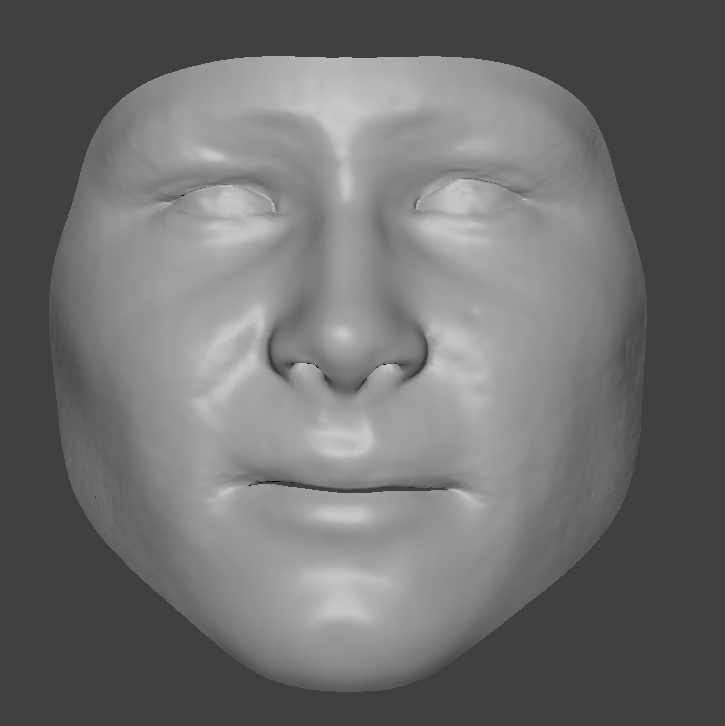}} &
\subfloat{\includegraphics[width = \figsz\textwidth]{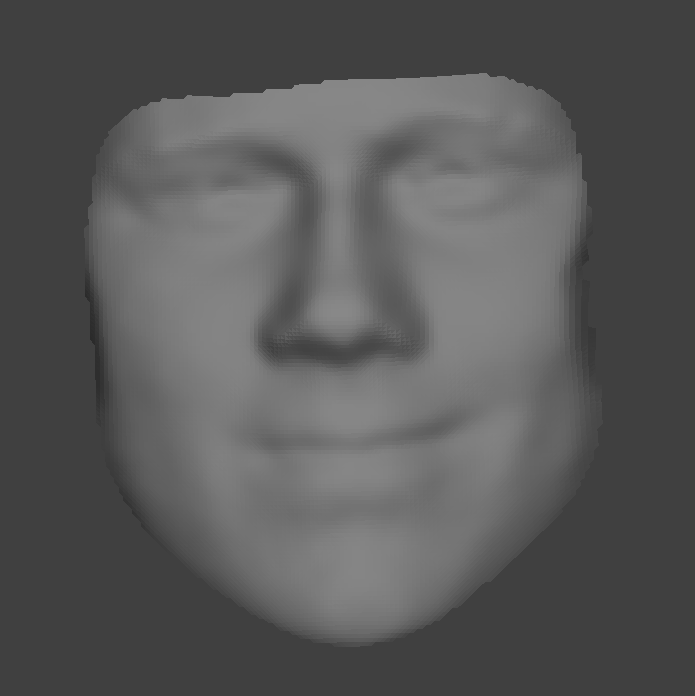}}
\end{tabular}
\caption{Qualitative results. Images are from BU4DFE-test. Our implementation
of fitting was used for the second column.}
\label{fig:qualitative}
\end{center}
\end{figure}

Table \ref{table:TewariFW} presents a comparison with Tewari et al. \cite{MOFA} and Garrido et al. \cite{GarridoRigs} 
on a selection of $9$ subjects from Face Warehouse \cite{FW} dataset. Version of Tewari et al. network with surrogate 
loss has been used for this and previous evaluations.
%
%
\setlength{\tabcolsep}{4pt}
\begin{table}
\begin{center}
\caption{Comparison with Tran et al. \cite{tran2017regressing} on MICC dataset \cite{MICC}. All numbers are in mm.}
\label{table:hassnerMicc}
\begin{tabular}{l|ll}
\hline\noalign{\smallskip}
Method & w.o. scale & w. scale \\
\noalign{\smallskip}
\hline
\noalign{\smallskip}
Tran \cite{tran2017regressing} & 1.83 $\pm$ 0.04 & 1.46 $\pm$ 0.03 \\
Our   & \textbf{1.70} $\pm$ 0.02 & \textbf{1.33} $\pm$ 0.02 \\
\end{tabular}
\end{center}
\end{table}
\setlength{\tabcolsep}{1.4pt}
%
%
%
%
\setlength{\tabcolsep}{4pt}
\begin{table}
\begin{center}
\caption{Comparisons on a selection from Face Warehouse dataset. All numbers are in mm.}
\label{table:TewariFW}
\begin{tabular}{l|lll}
\hline\noalign{\smallskip}
Method & Our & Tewari \cite{MOFA} & Garrido \cite{GarridoRigs}\\
\noalign{\smallskip}
\hline
\noalign{\smallskip}
Error & 1.8 $\pm$ 0.07& 1.7 & 1.4 \\
\hline
\end{tabular}
\end{center}
\end{table}
\setlength{\tabcolsep}{1.4pt}
\newline
\textbf{Qualitative Results.} Fig. \ref{fig:qualitative} shows a comparison with Jackson et al. \cite{VRN},
Tewari er al. \cite{MOFA} and Tran et al. \cite{tran2017extreme}  on a few images from BU4DFE-test.
%
%
\section{Conclusions}
%
%
We have presented an evaluation of monocular morphable model fitting algorithms and a learning framework. It
is demonstrated that incorporation of texture term into the energy function significantly improves fitting accuracy. 
Gains in the accuracy are quantified. We have trained a neural network using the outputs of the fitting algorithms 
as training data. Our trained network is shown to perform on a par with existing approaches for the 
task of monocular 3d face reconstruction while showing faster speed and smaller model size. Running time of our 
network on a mobile devise is shown to be $3.6$ milliseconds enabling real-time applications. Our datasets and code for 
evaluation are made publicly available. 
%
%
\setlength{\tabcolsep}{4pt}
\begin{table}
\begin{center}
\caption{Model size and running time comparison.}
\label{table:comparisons}
\begin{tabular}{llllll}
\hline\noalign{\smallskip}
Method   & Model size & GPU & CPU & ARM & AUC\\
\noalign{\smallskip}
\hline
\noalign{\smallskip}
Jackson \cite{VRN}              & 1.4 GB     & -          & 2.7 s  & -      & 0.849\\
Tewari \cite{MOFA}              & 0.27 GB    & 4 ms       & -      & -      & 0.848\\
Tran \cite{tran2017extreme}     & 0.35 GB    & 40 ms      & -      & -      & 0.810\\
Our                             & \textbf{1.5 MB}     & \textbf{1 ms}       & \textbf{1.8 ms} & 3.6 ms &  
\textbf{0.854}\\
\hline
\end{tabular}
\end{center}
\end{table}
\setlength{\tabcolsep}{1.4pt}
%
%
%
%
\bibliographystyle{splncs}
\bibliography{egbib}
\end{document}